\documentclass[runningheads]{llncs} 

\usepackage[english]{babel} 
\usepackage{amsmath,amssymb} 
\usepackage{graphicx} 
\usepackage{bm} 

\usepackage{hyperref} 
\usepackage{enumitem}
\usepackage{geometry}
\usepackage[numbers]{natbib} 
\usepackage{algorithm}
\usepackage{algpseudocode}  

\hypersetup{
    hidelinks,
    colorlinks=true,
    urlcolor=blue,
    citecolor=blue,
    linkcolor=blue
}
\usepackage{xcolor}
\usepackage{booktabs}
\definecolor{darkorange}{RGB}{255,140,0} 

\begin{document}

\title{MAB Optimizer for Estimating Math Question Difficulty via Inverse CV without NLP}

\author{
Surajit Das\inst{1}\orcidID{0009-0008-6692-6697}\thanks{The article is under review. Code used in this study may be provided upon reasonable request to the corresponding author} \and 
Gourav Roy\inst{2} \and 
Aleksei Eliseev\inst{3}\orcidID{0000-0002-4549-5356} \and 
Ram Kumar Rajendran\inst{4}\orcidID{0000-0002-0411-1782}
}

\institute{
ITMO University, St. Petersburg, Russia. 
\email{mr.surajitdas@gmail.com} \\
\and
Jadavpur University, Kolkata, India. 
\email{gouravroy2110@gmail.com} \\
\and
Moscow Institute of Physics and Technology, Moscow, Russia. 
\email{eliseev.av@mipt.ru} \\
\and
IIT Bombay, Mumbai, India. 
\email{ramkumar.rajendran@iitb.ac.in}
}

\maketitle

\begin{abstract}
The evolution of technology and education is driving the emergence of Intelligent \& Autonomous Tutoring Systems (IATS), where objective and domain-agnostic methods for determining question difficulty are essential. Traditional human labeling is subjective, and existing NLP-based approaches fail in symbolic domains like algebra. This study introduces the \textit{Approach of Passive Measures among Educands (APME)}, a reinforcement learning--based Multi-Armed Bandit (MAB) framework that estimates difficulty solely from solver performance data---marks obtained and time taken---without requiring linguistic features or expert labels. By leveraging the inverse coefficient of variation as a risk-adjusted metric, the model provides an explainable and scalable mechanism for adaptive assessment.

Empirical validation was conducted on three heterogeneous datasets: SKYBEN (middle school), TIMSS (international assessment), and IIT JEE Advanced (national entrance exam). Across these diverse contexts, the model achieved an \textbf{average $R^{2}$ of 0.9213} and an \textbf{average RMSE of 0.0584}, confirming its robustness, accuracy, and adaptability to different educational levels and assessment formats. Compared with baseline approaches---such as regression-based, NLP-driven, and IRT models---the proposed framework consistently outperformed alternatives, particularly in purely symbolic domains.

The findings highlight that (i) item heterogeneity strongly influences perceived difficulty, and (ii) variance in solver outcomes is as critical as mean performance for adaptive allocation. Pedagogically, the model aligns with Vygotsky’s Zone of Proximal Development by identifying tasks that balance challenge and attainability, supporting motivation while minimizing disengagement. This domain-agnostic, self-supervised approach advances difficulty tagging in IATS and can be extended beyond algebra wherever solver interaction data is available.

\end{abstract}


\section{Introduction} 

\addcontentsline{toc}{section}{Introduction} 

In the interdisciplinary framework of education and learning analytics, a futuristic research outcome is emergence of an effective Intelligent \& Autonomous Tutoring System (IATS). Hence developing an adaptive learning system for determining the difficulty-level \citep{alkhuzaey2023} of a question as a part of curriculum development is a significant and high priority research topic. Advancement of such autonomous and intelligent system is crucial in personalizing curricula, optimizing learning pathways, maintaining learner motivation and tailoring questions in real time based on learners’ dynamics governed by a complex dynamic system of different engagement variables (affective, cognitive, academic and Behavioural) along with the contextual variables associated with a learner. This system must be capacitated to estimate the ``expected payoff" or likelihood of a correct answer, while balancing risk factors such as failure, time taken, or learner demotivation \citep{kim2023}. Maintaining learners in the optimal learning zone—often described by Vygotsky’s Zone of Proximal Development (ZPD) \citep{murray2002toward,vygotsky1978mind} —is a must to prevent learners’ disengagement or panic/ phobia \citep{Ashcraft2002,15} in concerned subject.

So far, a few existing models attempt to reduce difficulty estimation to static, linguistic, or handcrafted features, limiting scalability and objectivity. Existing methodologies often require expert input or rely on NLP features such as syntax, embeddings, or problem type classifiers \citep{hong2021,lee2000,chen2005}. Besides, AI-enhanced educational systems are enriched with LLM or partially rule based approach to interpret the difficulty-level of a question in some cases, but, these approaches have several drawbacks including falling short in autonomously estimating question difficulty in symbolic / abstract domains like algebra \citep{walkington2019supporting,kinshuk2002}.
Traditional NLP-based models, which rely heavily on linguistic features, are invalid for pure algebraic questions that lack textual elements. Moreover, these systems are often tailored to homogenous learner populations and fail to scale across diverse cultural or educational backgrounds \citep{ram2024,brooks2021,roll2018}. In broader contexts such as MOOCs, where learners’ demographics and engagement patterns vary widely, lack of adaptability across subjects, context or learner population type is a pitfall for a personalized learning process \citep{zhu2021,liu2016,wang2007,elsabagh2021,mirata2020}. Also, this system is unable to dynamically tag a question with its difficulty-level out of a solver’s behavior (e.g., response time, accuracy) in real time without needing pre-labeled data or expert-curated features. Therefore, the significant gap that pertains to the development of autonomous, self-supervised, domain agnostic systems capable of estimating question difficulty based solely on performance metrics and without linguistic features or external labels.
This study aims to build a scientifically explainable, autonomous agent that classifies algebraic questions by difficulty based on solver behavior. The central research question is framed as \textbf{“Can an autonomous intelligent system identify the easiest question within a given set of algebraic questions, using only performance data like marks and time?”}

This study introduces the Approach of Passive Measures among Educands (APME), a self-supervised novel method for difficulty estimation grounded in Reinforcement Learning (RL) and the Multi-Armed Bandit (MAB) framework which requires no labeled data for training. APME is described as passive because: 1. It evaluates difficulty from solver behavior, not from the question creator’s assumptions. 2. It neither directly measure's a question’s static features grounded on NLP techniques nor the learner’s engagement which is passively captured through time domain. 3. It reduces feature dependence by using only marks obtained and time taken as input. The model learns difficulty rankings by optimizing performance-to-variance ratios, offering a domain agnostic mechanism that is scalable and explainable.

This study contributes to the field of intelligent tutoring systems and educational data mining in several novel ways:
\begin{itemize}
    \item A Passive, Reinforcement Learning-Based Model: We introduce APME, a domain-agnostic, self-supervised model that estimates question difficulty without relying on textual or structural features.
    \item Performance-Variance Trade-Off Mechanism: The model incorporates a statistically grounded difficulty estimation approach using the inverse of the coefficient of variation (mean/standard deviation), which balances expected solver performance with consistency.
    \item A conceptual framework distinguishing between intrinsic and extrinsic question characteristics and showing how both can influence difficulty perception.
    \item Domain Independence: APME is designed to work across domains by relying only on observable solver behaviors (marks and time), making it suitable for symbolic domains like algebra where NLP-based methods fail.
    \item Scalability for Large-Scale Adaptive Systems: The framework is computationally lightweight and interpretable, making it ideal for integration into real-time ITS and MOOC platforms.
    \item Empirical Validation Across Real Datasets: The methodology is tested using real student performance data collected from middle-school learners, and achieves high accuracy in estimating both difficulty means and variances with low RMSE values.
    \item Reusability and Reproducibility: Source code along with three diverse datasets is shared in a public repository as mentioned in the abstract.
\end{itemize}

Although the method is developed for algebra, APME’s design is domain-agnostic, capable of extending to other subjects as long as learner response metrics are available. Algebra is chosen due to the lack of existing robust methods for difficulty estimation in non-linguistic mathematical domains.

The remainder of this paper is structured as Section 2 (reviews related work), Section 3 (Academic Background \&  Theoretical Discussion), Section 4 (Methodology), Section 5 (Results \& Discussion) and Section 6 (Conclusion).


\section{Literature Review}
There exist a very few researches which pertain to “Finding the Difficulty Level of an algebric Question” from the year of 2000. All the approaches for “Finding the Difficulty Level of a Math Question” are largely based on text processing. Almost all the relevant works are mentioned below from  the year of 2000.\\

\cite{lee2000} developed an approach to measure difficulty-level of a problem in algebra through the concept of problem complexity. Their methodology identified six key complexity factors influencing problem difficulty: the number of perceived difficult steps, the number of steps required to solve a problem, numerical complexity, the number of logarithmic terms, the number of operations, and the degree of familiarity students had with the problem. These factors were combined into a multiple regression model to predict difficulty-level. This model allowed for preemptive difficulty estimation, bypassing the need for student testing, and was suitable for use in computer-assisted instructional systems. However, the total steps required to solve the problem often depend on the state of teaching-learning process; also, it is a relative matter subject to the discretion of paper setter/educator. The Author proposed that taken the shortest path of the solution is considered as the easy way out of the problem, however the shortest solution may make use a trick which itself may be difficult for an average student. In contrast the lengthy process maybe easier.  Moreover, while numerical complexities and logarithmic terms are definitely important characteristics for determining the difficulty level of a question there are many such lingos in algebraic questions, like, trigonometric functions, quadratic functions, complex formulas, etc. which are not covered. The question is how to determine the distance between the trigonometric function and logarithmic function scientifically if considering the logarithmic function as baseline. As the answer is unknown, this is found to be a significant gap of this research.\\

\cite{chen2005} proposed a personalized e-learning system based on Item Response Theory (IRT), termed PEL-IRT, to address learner disorientation and cognitive overload in web-based learning environments. Their methodology included modeling course material difficulty with the Rasch single-parameter logistic model and estimating learner ability using maximum likelihood estimation (MLE) based on explicit feedback. The system dynamically adjusted course difficulty using a collaborative voting mechanism that aggregated feedback from learners and experts. Course recommendations were personalized by aligning difficulty levels with the learner’s ability, facilitating adaptive learning pathways \citep{orsoni2023recommending}. In spite of this study the research has a serious drawback related to the voting approach mentioned. Normally expert and serious student is able to evaluate the difficulty level of a course material and this point is already addressed by the author. For an example, if a student skips some topic and doesn’t understand it well, the feedback provided by him against a course material related to that topic may deviate from the expected response with a very high probability. This situation will cause the system to capture noise; so, depending on the words of the teenager(s) for categorizing the questions based on the difficulty-level is not very promising. Furthermore, the views (perceptions) of the captured learners are not validated, despite their real-time performance of them. Also, the methodology is not appropriate for determining the difficulty of a question as it is dedicated to determine the difficulty level of a course material. \\

\cite{Segal2018} introduced Multi-Armed Bandits based Personalization for Learning Environments (MAPLE), a computational approach designed to personalize educational content for students, so that they can maximize their learning gains over time. MAPLE integrates difficulty ranking with a multi-armed bandit framework to optimize question selection. It estimated the expected learning gains for each question in a target set, and employs an exploration-exploitation strategy to dynamically select the next question for the student. Additionally, it also maintains and updates a personalized ranking of question difficulties in real-time, adapting to the student's progress as they engage with the content. The big gap of this research is the absence other characteristics of the dataset. Only expected gain does not ensure any confidence interval without z-score. \\ 

\cite{Jiao2023} developed a methodology for generating mathematical word problems (MWPs) with controlled difficulty levels. Their energy-based language model incorporated constraints on equations, vocabulary, and topics to ensure linguistic quality and creativity. However, this method heavily depended on NLP models such as GPT-2 and MAGNET, limiting its focus to linguistic aspects. The title of the work “Automatic Educational Question Generation with Difficulty Level Controls” indicates that the methodology measures the difficulty-level in some absolute scale and so generate it (MWP) but the approach is not a robust one to handle the cases of pure algebra which  reflects gap in our context of discussion. \\ 

Similarly, \cite{Ningsih2023} employed supervised learning techniques to classify the Indonesian and mathematical multiple-choice questions according to their difficulty levels. Their methods included Random Forest, Logistic Regression, Support Vector Machines (SVM), and Dense Neural Networks, which analyzed embeddings, lexical, and syntactic features.  This work is heavily dependent on the state of art of NLP (Natural Language Processing) which is not applicable in case of determining the difficulty-level of a pure algebraic question. The serious gap of this study is  a lack of treating the questions composed without natural language sentences (like pure algebraic questions). \\ 

\cite{kim2023} employed a two-step approach to analyze the difficulty-level of a mathematical question provided by ABLE Tech, a mathematical Learning Management System (LMS)-based solution platform targeting teenagers from elementary to high school. Initially, a t-test analysis was conducted to identify variables significantly correlated with difficulty, revealing that the correct answer rate, question type, and solution time had positive correlations with question difficulty. Subsequently, machine learning models, including logistic regression (LR), random forest (RF), k-Nearest Neighbor (k-NN) and extreme gradient boosting (xgboost), were utilized to classify questions based on their difficulty levels. Evaluation metrics such as accuracy, precision, recall, F1 score, AUC-ROC, Cohen’s Kappa, and Matthew’s correlation coefficient (MCC) were then applied to assess the models' performance. This approach is not a self-supervised approach, whereas, we have introduced the self-supervised learning which is much more prone to the ITS (Intelligent Tutoring System). \\

\cite{Shabana2023} introduced Unsupervised Skill Tagging (UST) to extract concept tags from student explanations to improve organization of instructional material. Data was collected during a linear algebra course where students provided problem-solving explanations. The text was pre-processed to remove irrelevant elements like punctuation and stop words, and frequent unigrams and bigrams were then extracted. These were matched with a predefined master tag list using exact word matching and Levenshtein distance, with a scoring function to determine tag relevance. Parameters were optimized using a train-test split, and the method's performance was evaluated against manual tagging using precision, recall, and F1 scores. UST aims to identify relevant concepts without supervision, leveraging student explanations to link questions to appropriate tags. Additionally, Linear Algebra course is not a generalized problem set pertaining to algebra whereas   by mentioning algebra we refer to mid school and high school algebra, which is beyond the scope of applying above research related to linear algebra. The study also uses linguistic analysis which is not relevant to the current topic of discussion, as these approaches fail to address any measuring technique required to determine the difficulty level of pure algebraic questions. \\ 

\cite{Fernandez-Fontelo2023} introduced a novel approach to predict question difficulty level using mouse movement features. Different measures, such as response time, travel distance, and directional changes (x and y flips), were calculated using the mousetrap package in R. Additional features, including initiation time, velocity, and acceleration, were explored for their potential to detect response difficulty. According to the authors, the study could not disentangle effects of the specific question from effects of the type of difficult, in addition the accuracies were only moderately high. Evidently, mouse movement  itself is not enough to describe the difficulty-level of any type of question in any subject field. \\

\cite{Nurhalimah2022} employed meta-analysis to measure the difficulty-level of mathematical problems. They examined student errors in solving mathematical problems based on Polya’s criteria, education levels, focus areas, and other moderator variables. Data were systematically collected from databases such as Scopus, DOAJ, WorldCat, Google Scholar, and the Garuda Portal, adhering to strict inclusion and exclusion criteria. The Preferred Reporting Items for Systematic Reviews and Meta-Analyses (PRISMA) framework guided the selection process through identification, screening, eligibility, and inclusion stages. Statistical data, such as the number of samples, error percentages across Polya criteria (understanding the problem, devising a plan, executing the plan, and reviewing), and effect size (ES) and standard error (SE) values, were prepared using Microsoft Excel and were analyzed using JASP software to produce results. In the above methodology, it is found that quantifying the degree of Understanding and precise Planning in order to solve a problem is hardly possible as these concepts are abstract and heavily reliant on human perception. Moreover, the authors have not provided any direction to address this issue which can be in consideration for the both autonomous and manual system. This is a significant research gap in respect of our research periphery. Consequently, this concept cannot be implemented to build an intelligent and autonomous system. \\

The above reflated studies mostly address various directions for determining the difficulty-level of a question, framed based on Natural language sentences. Hence, none of the methodologies used in the above studies are compatible for determining the difficulty level of an algebraic question.  Other studies also have  several limitations, which we already mentioned above, for measuring the difficulty-level of an algebraic question through Intelligent Tutoring System (ITS).
Hence,  we find a serious lack in existing research  of developing an intelligent and autonomous Tutoring systems for determining the difficulty level of an algebraic question without any intervention of human. Here we introduce a novel robust method, based on Multi-Armed Bandit Problems associated with Expected Value and Variance in order to measure the difficulty level of an algebraic problem which does not involve any sort of natural language sentence. The mean-variance trade-off is a must to depict distribution properly (the detailed discussion has been incorporated under section 3.6 named ``Coefficient of Variance"), which is a serious lack found in previous work.



\section{Academic Background \&  Theoretical Discussion}

Categorizing an algebraic question according to its level of difficulty—which is an intangible, relative, and multi-dimensional construct \citep{he2023question}—depends both on its subjective nature (intrinsic characteristics of the question) and on the solver's skills, which are influenced by contextual and engagement variables (extrinsic characteristics of the question).

\begin{itemize}
\item Intrinsic characteristics include static properties such as topological and taxonomical attributes, structural and linguistic complexity, conceptual depth, etc. \citep{albano2020,matsuda2019mathematical,sujay2024multi,chrysostomou2019analysing,anderson2014}, which remain invariant across different solvers.
    
\item Extrinsic characteristics depend mainly on the solver’s dynamics, which pertain to contextual variables (e.g., demographic factors) and several engagement variables—namely, cognitive, affective \citep{beymer2024students,payne2019}, behavioral, and academic factors. These variables govern the solver's receptive and productive skills and change over time.  
\end{itemize}

One way of developing an algorithm to measure the difficulty level of an algebraic question is through a direct approach: estimating all the above-mentioned characteristics along with the perspective of the question creator or instructor, who would rely on their experience \citep{hong2021,lee2000,chen2005}.

An alternative method introduced in this study is the Approach of Passive Measures among Educands (APME), inspired by the Multi-Armed Bandit problem and addressed using Reinforcement Learning. As mentioned in the introduction, we capture the effects of engagement variables on learning outcomes through their time-domain dynamics, rather than by directly measuring these variables. This is considered a passive measure.

\subsection{Notion of Multi-Armed Bandid (MAB) Framework:} 
The MAB problem models sequential decision-making under uncertainty, where an agent selects actions (``arms") with unknown reward distributions to maximize cumulative gains \citep{Vermorel2005}. It balances exploration (trying new arms) and exploitation (choosing optimal arms) to minimize regret \citep{Audibert2009}. In this study, each problem acts as an arm, with rewards derived from solver performance (marks/time), enabling adaptive difficulty estimation.

\subsection{Theoretical Justification for Coefficient of Variation in MAB}
\label{sec:cv-theory}

The coefficient of variation (CV), defined as $\text{CV} = \sigma/\mu$, serves as a critical metric for risk-sensitive decision-making in the Multi-Armed Bandit (MAB) framework \cite{Audibert2009}. Below, we formalize its advantages over alternative metrics and its pedagogical implications.

\subsubsection*{Scale-Invariance and Regret Bounds}
CV's scale-invariance property enables comparison across diverse datasets (e.g., middle-school algebra, JEE Advanced, etc.) by normalizing risk ($\sigma$) relative to expected gain ($\mu$). This aligns with MAB's goal of minimizing \textit{cumulative regret} $R_T$:

\begin{equation}
R_T = T \cdot \mu^* - \sum_{t=1}^T \mu_{a_t},
\end{equation}

where $\mu^*$ is the optimal arm's mean. By prioritizing arms with high $\text{CV}^{-1} = \mu/\sigma$, the agent tightens regret bounds, as $\sigma \propto \sqrt{\mathbb{E}[R_T]}$ \cite{Vermorel2005}. For Gaussian rewards, the probability of achieving $\mu \pm \epsilon$ is:

\begin{equation}
P(|\text{Gain} - \mu| \leq \epsilon) \approx \text{erf}\left(\frac{\epsilon \sqrt{2}}{\sigma}\right),
\end{equation}

where lower $\sigma$ (thus higher $\text{CV}^{-1}$) increases confidence (e.g., $\sigma_1 = 0.34$ yields 22.8\% probability vs. $\sigma_2 = 1.2$ at 6.7\% for $\epsilon = 0.1$).

\subsubsection*{Comparison to Alternative Metrics}
Table~\ref{tab:risk-metrics} contrasts CV with common risk-adjusted measures. CV is preferred for its interpretability and absence of external benchmarks (e.g., Sharpe ratio's risk-free rate).

\begin{table}[htbp]
\centering
\caption{Comparison of risk metrics for difficulty estimation.}
\label{tab:risk-metrics}
\begin{tabular}{p{3cm}p{6cm}p{6cm}}
\toprule
\textbf{Metric} & \textbf{Advantages} & \textbf{Limitations} \\
\midrule
CV & 
Scale-invariant; intuitive for educators & 
Unstable for $\mu \approx 0$ \\

Sharpe Ratio & 
Standard in finance & 
Requires risk-free baseline \\

Gini Coefficient & 
Robust to outliers & 
Computationally expensive \\
\bottomrule
\end{tabular}
\end{table}

\subsubsection*{Pedagogical Alignment with ZPD}
\begin{figure}[htbp]
    \centering
    \begin{minipage}{0.66\textwidth}
        \centering
        \includegraphics[width=\linewidth]{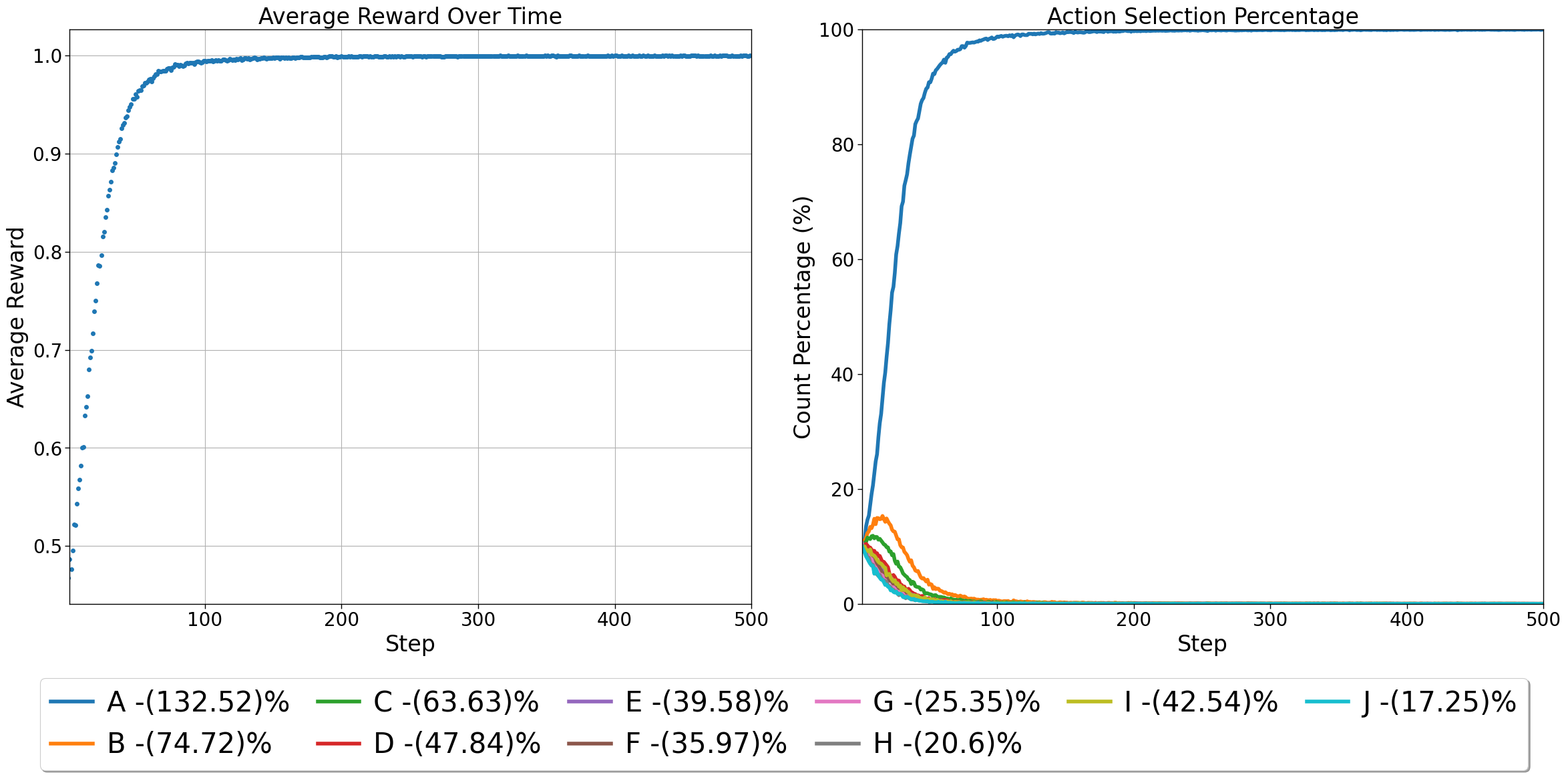}
    \end{minipage}
    \hfill
    \begin{minipage}{0.32\textwidth}
        \centering
        \includegraphics[width=\linewidth]{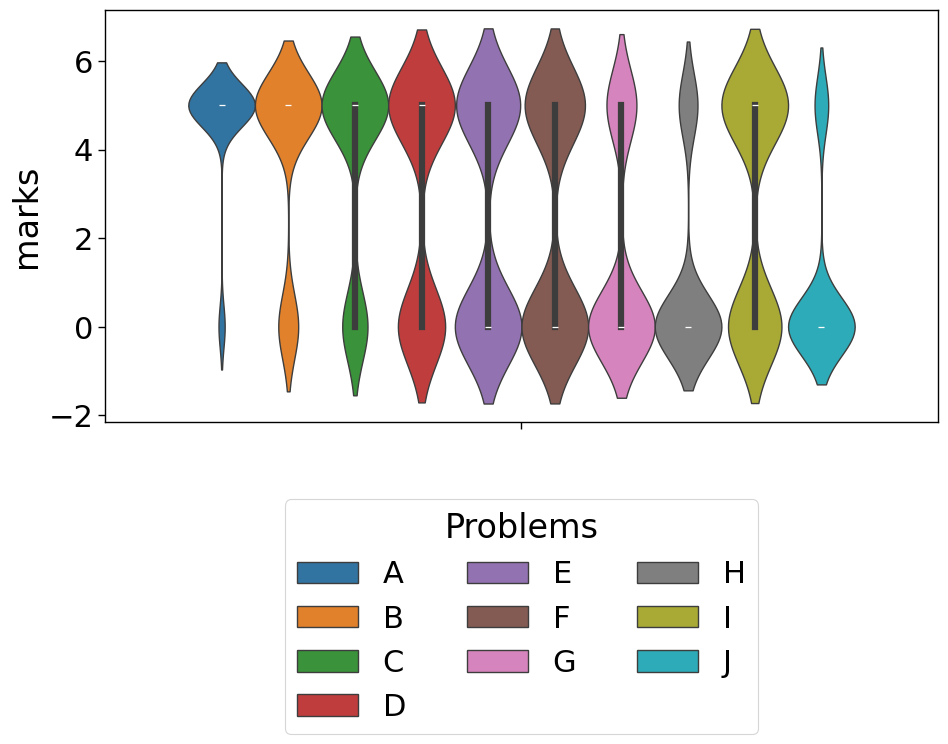}
    \end{minipage}
    \caption{Analysis of the SKYBEN Dataset: (Left) Average Reward Over Time. (Center) Cumulative count of arm selections Over Time. (Right) Distribution of marks.}
    \label{fig:skyben}
\end{figure}

CV maps to Vygotsky's Zone of Proximal Development (ZPD) by quantifying \textit{predictable challenge}:
\begin{itemize}
    \item \textbf{Low CV}: High consistency but potentially trivial ($\mu \ll$ solver ability).
    \item \textbf{High CV}: Unreliable outcomes, risking frustration \cite{Ashcraft2002}.
\end{itemize}
Empirically, problems with moderate CV (e.g., problem B in Fig.~\ref{fig:skyben}) kept learners engaged longest.

\subsection{Pseudo Value of a Question}
The pseudo value of a question refers to the assigned numerical marks awarded based on a predefined scoring rubric, typically used for practical evaluation purposes. These marks act as a proxy—though not equivalent—for the intrinsic worth of the question and are determined by educators according to perceived difficulty, expected length of the solution, or comparative assessment against other items in the same test. However, this assignment is inherently subjective, often based on heuristic judgments rather than empirically measured performance data. Consequently, pseudo values offer only a relative estimation of question difficulty and are susceptible to biases stemming from inconsistent marking schemes or domain-specific norms. Within the reinforcement learning framework adopted in this study, pseudo values function as inputs to model student gain and behavior but do not serve as the final arbiter of question difficulty.

\subsection{True Value of a Question:}
In contrast, the true value of a question is a fixed, invariant property representing its intrinsic complexity, independent of solver characteristics or contextual factors. It encapsulates the essential conceptual depth, structural complexity, and logical demands embedded in the question itself. Unlike pseudo values, which vary with human perception and contextual framing, the true value remains constant across solvers and settings. However, this value is abstract and not directly observable; it must be inferred indirectly through aggregated performance metrics such as solver accuracy, time invested, and consistency of outcomes. In the proposed reinforcement learning–based model, the true value is approximated through expected gain adjusted by the coefficient of variation, enabling a data-driven and domain-agnostic estimation of difficulty that aligns more closely with an objective understanding of question challenge.

\subsection{Modeling the Problem:}
A question's challenge manifests through multiple dimensions of complexity, including Syntactic Complexity (e.g., linguistic structure in word problems), Topological Complexity (e.g., network dependencies among concepts), Conceptual Complexity (number and abstraction of underlying principles), Procedural Complexity (number and nature of solution steps), Contextual Complexity (required real-world knowledge), etc. These components form a feature vector $\mathbf{c} = (c_1, c_2, \ldots, c_n)$, representing the question's invariant and inherent intrinsic complexity.

This intrinsic complexity necessitates a certain effort investment, which is mediated through the solver's engagement variables ($e_{\text{cog}}, e_{\text{beh}}, e_{\text{aff}}, \ldots$) and other contextual factors. If $\mathbf{e}$ denotes a vector combining these engagement and contextual variables, the entire system can be synthesized into a single latent scalar $p$. This scalar $p$ represents the true value of the question—defined as the objective effort investment required for its solution, also termed the realized effort. This is formalized by the function $\mathcal{G}$:

\begin{equation}
p = \mathcal{G}(\mathbf{c}, \mathbf{e})
\end{equation}

where $\mathcal{G}(\cdot)$ is a symbolic function mapping the complexity features and engagement profile to the required cognitive investment. The final perceived difficulty score $\zeta$ is generated by the composition of functions:

\begin{equation}
\zeta = \mathcal{D} \circ \mathcal{G}
\end{equation}

This function $\zeta$ can be a simple composition or a more complex functional mapping.

As the true value $p$ and engagement vector $\mathbf{e}$ are latent, we employ a strategy of relative measurement. The framework proxies these latent variables through observable interactions: specifically, the invested time (a proxy for effort) and the assigned score. When a solver invests effort (proxied by time $t$) to solve a question, the interaction yields a return. This return consists of two components: a reward $R$, which constitutes an externally assigned performance-based incentive defined by the assessment policy, intrinsic cognitive enrichment, which recuperates the invested effort and is quantified by the true value $p$ (the intrinsic cognitive investment). Thus, for any problem, the return ($\mathcal{S}$) is modeled as:

\begin{equation}
\mathcal{S} = R + p
\end{equation}

$ 
\text{In this model, Performance is defined as } \eta:= \alpha \frac{\text{$mark$}}{\text{$time$}} = \frac{v}{t} \quad  \text{where $mark$} \in \{0, 1\}$. Here, $v$ is the pseudo-value  (the externally assigned marks), an exogenous value assigned based on correctness and scaled by an anthropogenic scalar $\alpha$. This scalar incorporates external assessment policies set by the paper-setter (e.g., weighting for question importance). If $\alpha = 1$, the reward is unbiased; otherwise, it introduces a policy-based scaling.  
 The average performance for $i$th problem is calculated as: \begin{equation}
\bar{\eta_i} =  \frac{1}{k_i} \cdot \sum_{m=1}^{k_i} \frac{v_i}{t_i^{(m)}}
\end{equation}

where $v_i$ denotes the pseudo-value assigned to problem $i$, $t_i^{(m)}$ represents the time taken in the $m^{\text{th}}$ instance of all responses to problem $i$.

If a solver invests excessive time to solve the problem, perfomance decreases; $t \to \infty \implies \eta \to 0$, indicating low efficiency and thus maximal perceived difficulty. This formulation captures the inverse relationship between observed performance efficiency and the underlying perceived difficulty.

As mentioned earlier, a reward $R$, which constitutes
an externally assigned performance-based incentive as the assessment policy is defined as 

\[
R = \frac{\mathbb{E}[\eta]}{\sqrt{\mathrm{Var}(\eta)}} = \frac{\bar{\eta}}{\sigma_{\bar{\eta}}}
\]

\begin{equation}
\mathcal{S}_i =  \frac{1}{\sqrt{\mathrm{Var}(\eta_i)}
k_i} \cdot \sum_{m=1}^{k_i} \frac{v_i}{t_i^{(m)}} + p_i
\end{equation}

$v_i$ denotes the pseudo-value assigned to problem $i$, $t_i^{(m)}$ represents the time taken in the $m^{\text{th}}$ instance of all responses to problem $i$  $p_i$ is the true value associated with ith problem (class) $i$. $\mathcal{S}_i$  is the average return of from $i$th problem.

Hence, solving a question is equivalent to pulling an arm and observing a stochastic reward which is grounded to the tangent of the performance–risk angle. To guide arm selection in the MAB framework, the agent uses this performance-to-risk ratio. The formulation prefers arms with high expected performance and low variance (i.e., stable easiness). It serves as the objective for our bandit optimization: this ratio acts as the agent’s reward signal and is optimized through exploration and exploitation. Higher values indicate easier problems with lower variability in solver outcomes.

\subsection{Introduction to Modulator} 

In reality, the marking system is not only restricted to anthropogenic scalars but is also extended to various policies (such as negative marking), which can be influenced by solver behavior. For an instance, a solver may prolong time by sitting idle to minimize penalties in the case of negative marking, or may attempt an answer based on an educated guess.  To handle such uncertainties, we introduce weighting vector called the \textit{Modulator}, defined as  
\[
A =
\begin{bmatrix}
A_1 \\
A_2
\end{bmatrix},
\]  
such that, for any problem the return  \( \mathcal{S} \) is reformulated as  
\[
\mathcal{S'} = A^{\top} M,
\]  
where: \[ M^{\top} = [R, p ], \]

Hence, for $i$th problem, the equation (5) can be re-written as: 
\begin{equation}
\mathcal{S'}_i =  A_1 \cdot R_i + A_2 \cdot p_i
\end{equation}

Using this Modulator, we reframe the reward without loss of generality to address the aforementioned scenarios, as demonstrated with the JEE-Advanced dataset. 

While introduced here as a theoretical component, the modulator 
$A$ is calibrated empirically for specific assessment policies (e.g., negative marking) as detailed in Section 4.5, ensuring the model's adaptability to diverse real-world scoring systems.


\section{Methodology}
\label{sec:methodology}

This section outlines the data collection, preprocessing steps, problem formulation, reward structure, algorithm design, and simulation setup used to estimate the difficulty of algebraic questions through a reinforcement learning-based multi-armed bandit (MAB) approach considering problem as an arm.

\subsection{Data Acquisition}
\label{subsec:data_acquisition}

Three datasets were utilized to evaluate the proposed model:

\begin{enumerate}
\item \textbf{Skyben Dataset:} Contains performance records of 200 Indian middle-school students (Grade VII) collected during 2019--2021 through online assessments. Each record includes student ID, problems (Problem ID), time taken (in milliseconds), and marks (5 for correct answer while a wrong answer secures zero). The problem set was deliberately designed to avoid guided or correlated sequences of questions, thereby reducing learning bias across problems.
\item \textbf{TIMSS Dataset:\footnote{\url{https://sites.google.com/site/assistmentsdata/home}, \url{https://sites.google.com/site/assistmentsdata/datasets/2012-13-school-data-with-affect}}} The study utilized the publicly available TIMSS (Trends in International Mathematics and Science Study, 2012-13 School Data with Affect) dataset, which provides student performance data across mathematics and science domains. The focus was specifically on mathematics-related responses, as they provide a reliable proxy for quantifying varying difficulty levels across questions. The original dataset is huge and diverse. We extracted a subset of 24 questions, which had over 89,000 responses with associated time tracking data.
\item \textbf{JEE-Advanced Dataset:\footnote{\url{https://jeeadv.ac.in/reports/2024.pdf}, pg.~24}} 
Constructed from publicly available response statistics of the JEE-Advanced 2024 exam. It includes question-wise correct, incorrect, and unattempted response counts from 180,200 candidates. 

The dataset also included multi-correct questions which involved partial marking schemes. However, in the data provided in JEE-Advanced reports, the distribution of these partially correct marks (i.e., +1, +2, etc.) is not available. Therefore, such questions (here Q5, Q6, and Q7) could not be used in the simulation and have been excluded.

For the questions selected, the marking scheme is as follows:
\begin{itemize}
    \item Q1--Q4, Q14--Q17: +3 for correct, $-1$ for incorrect
    \item Q8--Q13: +4 for correct, 0 for incorrect
\end{itemize}

\end{enumerate}

Each dataset records solver behavior in terms of problem-solving time and accuracy, making them suitable for performance-based difficulty estimation.

\subsection{Data Preprocessing}
\label{subsec:data_preprocessing}

The acquired data underwent several preprocessing steps to ensure consistency and suitability for the simulation framework. Unwanted columns were removed. Subsequently, missing entries, ambiguous responses, and incomplete records were excluded. The data was then filtered to retain only complete sets of responses across the selected mathematical items. This preprocessing step ensured that each learner's record contributed consistently to the modeling process, avoiding biases caused by uneven participation or partially filled response sets.

\subsubsection{Usecase: Data preprocessing for TIMSS}
\label{subsubsec:timss_preprocessing}

The raw TIMSS dataset has the shape of (6123270, 35). It contained information such as problem identifiers, timestamps marking the start and end of each attempt, and whether the response was correct. We extracted a subset of the records where each problem had a minimum of 3500 recoded perfomances, and the shape was reduced to (89699, 35) with 24 problems.

Following this, the temporal features were cleaned and standardized. The start time and end time columns were converted into proper datetime objects, and any rows containing invalid or missing timestamps were removed to preserve consistency. From these cleaned timestamps, a new feature was engineered to capture the duration of each attempt. The time taken was calculated as the difference between the end and start times, expressed in milliseconds and stored as an integer for precision.

The final dataset was then constructed by retaining only the essential variables required for further analysis: the problem identifier, start time, end time, time taken per attempt, and correctness of the response. To ensure direct compatibility with the simulation code, the \texttt{time taken} column was renamed to \texttt{milsec} and the \texttt{correct} column was renamed to \texttt{marks}.

\subsection{True Target Variable Determination, Aggregation \& Normalization}
\label{sec:target_var}

To maintain clarity in terminology, we distinguish between the instantaneous performance per attempt ($\eta$) and the aggregate metrics derived from it.
For each problem, the mean ($\mu_{\eta}$) and standard deviation ($\sigma_{\eta}$) of the solver's \textbf{instantaneous performance} $\eta$ (as defined in Eq.~6) were computed. A \textbf{Derived Performance} metric ($\psi$) was then defined as the inverse coefficient of variation:

\[
\psi = \frac{\mu_{\eta}}{\sigma_{\eta}}, \tag{8}
\]

which served as the \textbf{true reward signal} $R$ (as theorized in Section~3.6), reflecting the stability and risk-adjusted efficiency of solver outcomes. Higher values of $\psi$ indicated problems that were consistently solved with high efficiency and low variability, while lower values corresponded to higher uncertainty and perceived difficulty. This ground-truth measure remained hidden from the agent and was used only to validate experimental outcomes.

In parallel, learners' performances were aggregated into a success-to-failure ratio, modeling each problem as a stochastic arm with an underlying probability of success and forming the empirical basis for simulating learner--question interactions. To address unequal participation, success ratios were normalized by total attempts to yield comparable difficulty metrics, followed by min--max scaling to $[0, 1]$, ensuring uniformity and stability in subsequent optimization.

\subsection{Probability Assignment to Simulated Environment}
Normalized metrics were used to assign probabilities in the simulated bandit environment, where each problem’s success rate represented the hidden reward probability of its arm. Easier questions corresponded to higher probabilities of success and harder ones to lower, yielding a probabilistic representation of the difficulty distribution and enabling faithful simulation of student–problem interactions.  

\begin{equation}
p_i = \frac{\text{Derived Performance}_i}{\sum_k \text{Derived Performance}_k},
\end{equation}
These probabilities governed the likelihood of each problem being selected during simulation and introduced weighting schemes to balance exploration across varying difficulty levels.

\subsection{Modulator Determination} 
In the TIMSS and SKYBEN datasets, the marking scheme was simplified by excluding a negative marking strategy. SKYBEN uses $\alpha = 4$ to construct a pseudo-value (a correct answer receives 4 points), while TIMSS uses $\alpha = 1$ (a correct answer receives 1 point). In both cases, we plug $A_1 = 1$ and $A_2 = 0$ into our modulator. This anchors the true value of a question to a 0 reference, keeping the reward calculation unaltered.

In the case of the JEE Advanced dataset, the presence of a negative marking scheme (where 1 point is deducted for every wrong answer) for some questions requires us to set $A_1 = 1$ and $A_2 = \frac{1}{\text{true-value}}$. This affine transformation effectively shifts the entire scoring system into the positive domain. The operation preserves the ordinal relationships between outcomes and maintains the linear structure of the original scoring scheme, making it suitable for subsequent analysis without loss of generality.

\subsection{Bandit Simulation}

As stated earlier, each problem was modeled as an arm in a Multi-Armed Bandit (MAB) framework. The agent iteratively interacted with the environment by selecting arms and receiving stochastic rewards generated from the derived performance distribution. To balance exploration and exploitation, Thompson Sampling was implemented.

\subsection{Evaluation Metrics}

The performance of all bandit algorithms was assessed using the coefficient of determination ($R^2$) and the Root Mean Square Error (RMSE).  The $R^2$ score measured how well the predicted rewards aligned with observed solver outcomes:
where $y_i$ are the observed values, $\hat{y}_i$ are the predicted values, and $\bar{y}$ is the mean of the observed values. The RMSE quantified the deviation between predicted and actual values:
Together, these metrics provided a comprehensive evaluation of the predictive validity and stability of the bandit-based difficulty interpretation framework.

\section{Results and Discussion:}

We analyzed three experimental datasets spanning heterogeneous contexts and scales: 
Skyben (10 problems; 2,000 trials), TIMSS (24 problems; 89,699 records), and IIT JEE Advanced 
(14 problems; 2,522,800 records). The computational framework discussed previously used in all three analyses.

\begin{table}[htbp]
\centering
\caption{Performance Evaluation of Multi-Armed Bandit Model on Different Datasets}
\label{tab:mab_evaluation}
\begin{tabular}{lcccc}
\toprule
\textbf{Dataset} & \textbf{R² Score} & \textbf{RMSE} & \textbf{Steps} & \textbf{Experiments} \\
\midrule
SKYBEN Dataset & 0.8011 & 0.1446 & 5,000 & 10,000 \\
TIMSS Dataset & 0.9845 & 0.0130 & 10,000 & 240,000 \\
JEE Advanced Dataset & 0.9784 & 0.0175 & 3,000 & 70,000 \\
\bottomrule
\end{tabular}
\end{table}

\begin{figure}[htbp]
\centering
\begin{minipage}{1\textwidth}
    \centering
    \includegraphics[width=\textwidth]{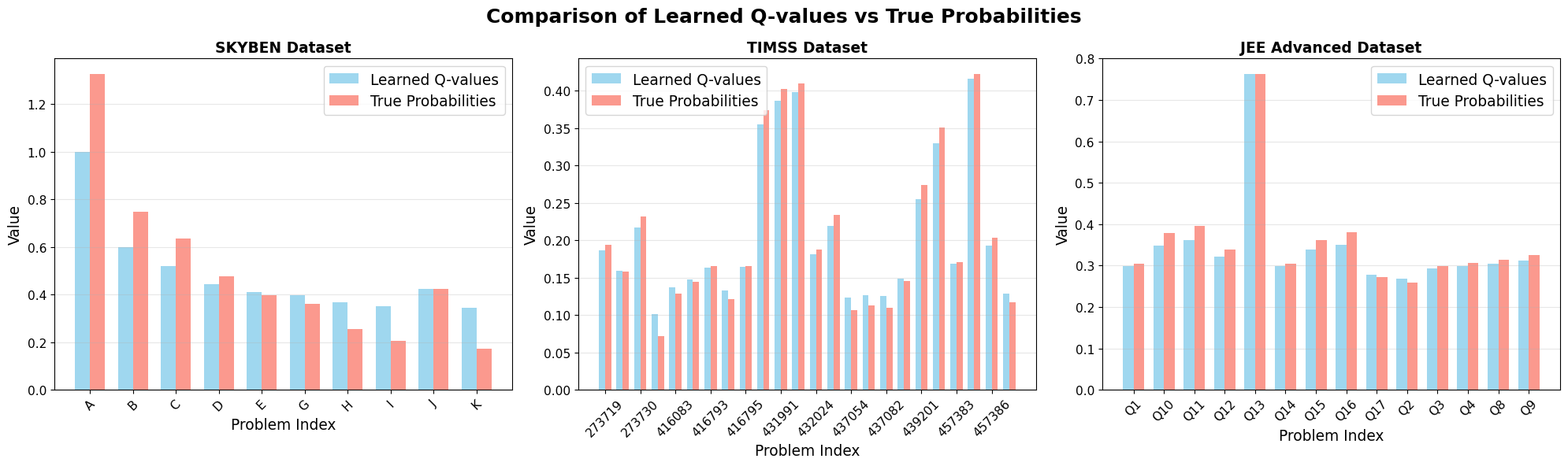}
    \caption{Normalized mean "derived-performance" for highlighted top/bottom problems within each dataset (Data is scaled by the 5 times for distinguishable visibility).}
    \label{fig:top_bottom_normalized}
\end{minipage}
\end{figure}

Problem-wise analyses revealed consistent gradients in difficulty. 
In the Skyben dataset, category A achieved the highest mean performance 
($\mu \approx 6.94$, $\sigma \approx 2.16$) while category J was at the lower extreme 
($\mu \approx 0.49$, $\sigma \approx 1.17$). 
TIMSS showed similar spread, with items 416796 ($\mu \approx 7.59$, $\sigma \approx 6.89$) 
and 457384 ($\mu \approx 7.32$, $\sigma \approx 5.89$) outperforming item 437054 
($\mu \approx 0.49$, $\sigma \approx 1.55$). 
IIT JEE Advanced exhibited much larger absolute performance values due to its 
scoring and timing scale, with Q16 ($\mu \approx 237,305$, $\sigma \approx 172,298$) 
and Q12 ($\mu \approx 229,048$, $\sigma \approx 186,992$) outperforming Q13 
($\mu \approx 103,541$, $\sigma \approx 37,466$). 
Highlights are provided in Table~\ref{tab:problem_highlights}. 
Fig.~\ref{fig:top_bottom_normalized} represents comparison of learned data and actual probability.

\begin{figure}[htbp]
    \centering
    \begin{minipage}{0.66\textwidth}
        \centering
        \includegraphics[width=\linewidth]{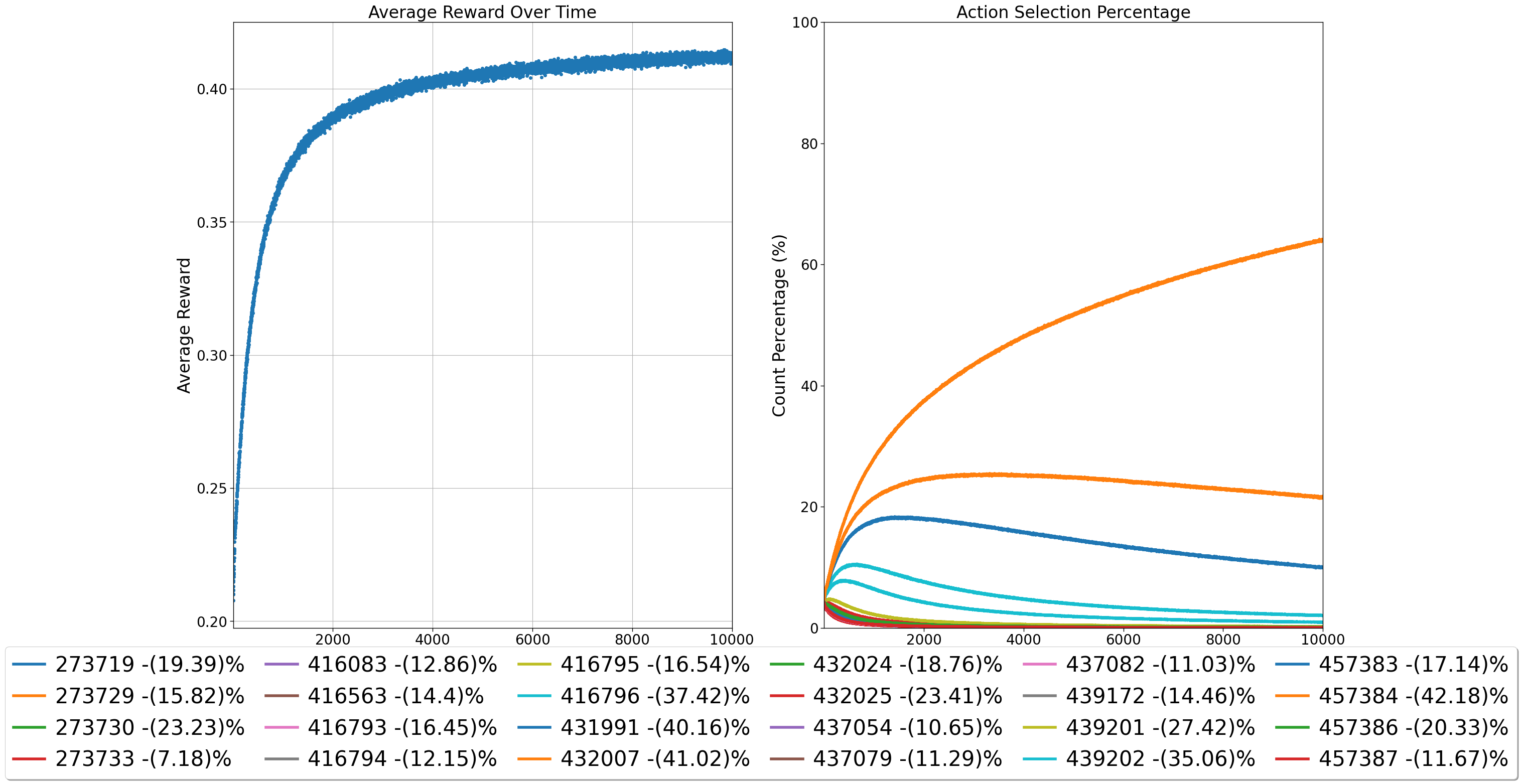}
    \end{minipage}
    \hfill
    \begin{minipage}{0.32\textwidth}
        \centering
        \includegraphics[width=\linewidth]{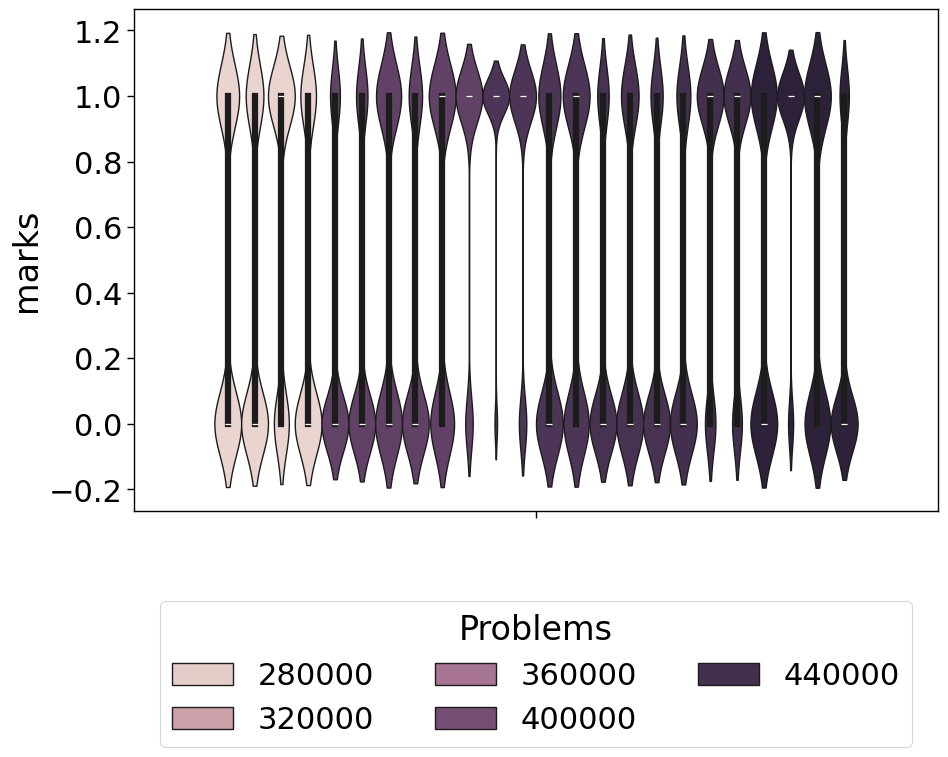}
    \end{minipage}
    \caption{Analysis of the TIMSS Dataset: (Left) Average Reward Over Time. (Center) Cumulative count of arm selections Over Time. (Right) Distribution of marks.}
    \label{fig:TIMSS}
\end{figure}

\begin{figure}[htbp]
    \centering
    \begin{minipage}{0.64\textwidth}
        \centering
        \includegraphics[width=\linewidth]{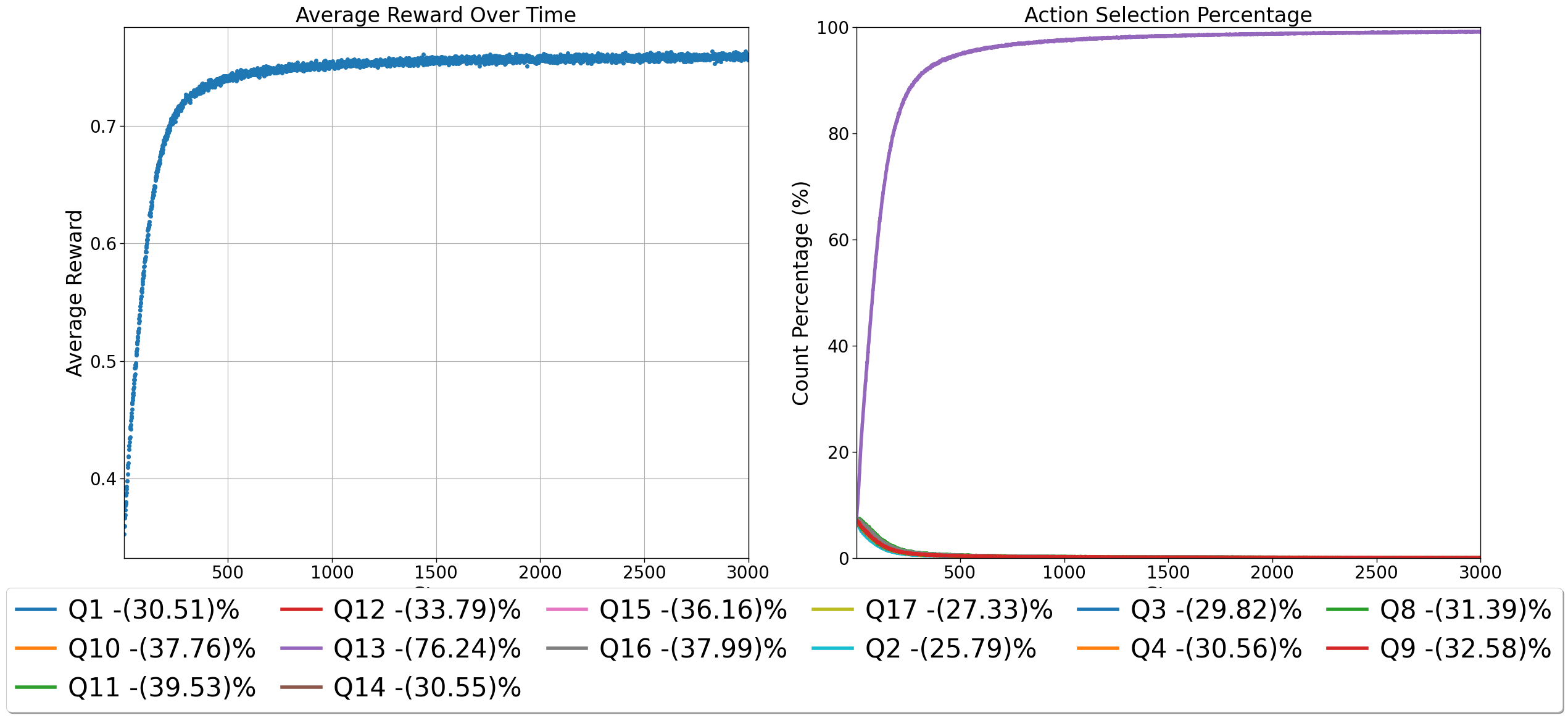}
    \end{minipage}
    \hfill
    \begin{minipage}{0.32\textwidth}
        \centering
        \includegraphics[width=\linewidth]{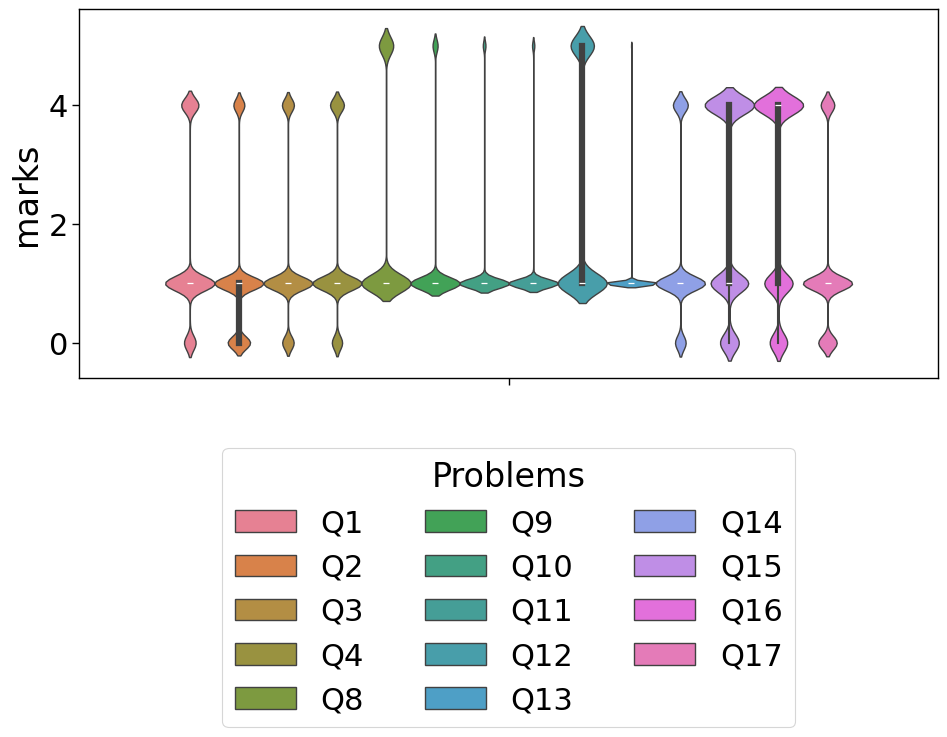}
    \end{minipage}
    \caption{Analysis of the JEE Advanced Dataset: (Left) Average Reward Over Time. (Center) Cumulative count of arm selections Over Time. (Right) Distribution of marks.}
    \label{fig:iit_jee_results}
\end{figure}

Comparison with Baseline Approaches establishes that the proposed model significantly outperforms traditional methods in estimating question difficulty, as summarized in Table~\ref{tab:model_comparison}. While previous models relied on linguistic features, expert annotations, or heuristic rule-based strategies, they lacked adaptability to purely symbolic or algebraic question formats and often failed to generalize across domains. The RL model, in contrast, is entirely self-supervised and domain-agnostic, requiring only solver performance data (marks and time) as input. It avoids the pitfalls of subjective tagging and language-based feature extraction.

Models such as GPT-2-based generators and mouse-movement trackers either focus on natural language or engagement proxies, offering limited insight into question difficulty—especially for algebraic content that lacks semantic structure.

\begin{table}[htbp]
\centering
\caption{Some problem-wise highlights showing the best- and worst-performing items in each dataset. Mean performance is normalized; Std denotes within-item variability.}
\label{tab:problem_highlights}
\begin{tabular}{lcccc}
\hline
Dataset & Problem & Mean Performance & Std & Rank \\
\hline
Skyben & A & 6.94 & 2.16 & Top \\
Skyben & C & 4.39 & 2.84 & Top \\
Skyben & H & 0.71 & 1.42 & Bottom \\
Skyben & J & 0.49 & 1.17 & Bottom \\
TIMSS & 416796 & 7.59 & 6.89 & Top \\
TIMSS & 457384 & 7.32 & 5.89 & Top \\
TIMSS & 437054 & 0.49 & 1.55 & Bottom \\
IIT JEE Advanced & Q16 & 237,305 & 172,298 & Top \\
IIT JEE Advanced & Q12 & 229,048 & 186,992 & Top \\
IIT JEE Advanced & Q13 & 103,541 & 37,466 & Bottom \\
\hline
\end{tabular}
\end{table}

\begin{table}[h!]
\centering
\caption{Comparison of our model with existing approaches.}
\label{tab:model_comparison}
\begin{tabular}{l l l l}
\toprule
\textbf{Model} & \textbf{Metric} & \textbf{Score} & \textbf{Remarks} \\
\midrule
Our Model & RMSE, $R^2$ (Avg. for 3 datasets) & $0.0584 $, $0.9213$ & Continuous, domain-agnostic \\
Lee \& Heyworth & RMSE, $R^2$ & 0.65 & Based on rule-based regression \\
NLP + RF & F1 Score & 0.78 & Text-dependent, not symbolic \\
Item Response Theory & Correlation & High & Expert-vote bias, lacks validation \\
MAPLE & Arm accuracy & N/A & Decision-focused, no scoring \\
GPT-2 + MAGNET & N/A & N/A & Language model, irrelevant for algebra \\
Mouse Movement & Engagement Metric & Moderate & Indirect proxy, not scalable \\
\bottomrule
\end{tabular}
\end{table}

Across all three settings, performance was strongly item-dependent. 
The Skyben and TIMSS datasets---both dichotomously scored---expose clear 
separations between tractable and challenging problems, with wider within-item 
variance in higher-performing categories. 
The IIT JEE dataset, by virtue of scale and partial credit, provides a more 
graduated view: high-mean items (Q12, Q16) also display substantial variance, 
implying that they differentiate ability well but may be sensitive to 
strategy selection and time management.

Two practical conclusions emerge. First, \textbf{item heterogeneity is a 
dominant driver of observed efficiency}, regardless of domain. 
Second, \textbf{variance matters}: high means paired with large standard 
deviations suggest opportunities for targeted scaffolding or adaptive allocation 
rather than uniform practice.

Methodologically, these findings motivate the adoption of 
\textit{adaptive selection schemes} (e.g., Thompson sampling or other 
multi-armed bandit methods) to balance exploration of difficult items 
with exploitation of items that currently yield efficient gains. 
In high-stakes, large-scale contexts such as IIT JEE, partial-credit 
dynamics further argue for \textit{time-aware reward shaping}, since 
incremental improvements in marks per unit time can compound meaningfully 
over the course of assessment or practice.

The findings underscore the potential of passive, performance-based analytics in educational systems. By capturing solver interaction data and processing it through a reinforcement learning framework, the system aligns closely with Vygotsky’s Zone of Proximal Development (ZPD), identifying tasks that are neither too easy nor too difficult. This contributes to optimal challenge zones, promoting sustained motivation and reducing the risk of disengagement or anxiety.

Importantly, the model's use of slope-based classification facilitates dynamic curriculum design. Educators or intelligent tutoring systems (ITS) can assign problems based on both difficulty and risk preference, enabling fine-grained personalization.

For example, a student exhibiting high consistency but moderate performance may benefit more from questions like Problem B (low variance) from SKYBEN Dataset, while more confident learners can be progressively challenged with problems like Problem C or D, which offer higher gain but also greater variability.

\section{Conclusion}

This study presents a domain-agnostic, self-supervised reinforcement learning framework for estimating the difficulty of algebraic questions using only performance data (marks and time taken). Leveraging the Multi-Armed Bandit paradigm and optimizing the inverse coefficient of variation, the model effectively ranks problem difficulty without relying on NLP, expert annotation, or pre-labeled datasets. Experiments across datasets ranging from middle school assessments to national entrance exams demonstrated low RMSE values, confirming the framework’s robustness, accuracy, and adaptability.

\textbf{Practical significance:} The model is lightweight, explainable, and scalable, suitable for real-time deployment in adaptive learning platforms, including Intelligent and Autonomous Tutoring Systems (IATS). Its performance-centric design allows seamless integration into educational technologies, particularly in symbolic domains like algebra where traditional NLP-based methods are limited.

\textbf{Pedagogical significance:} The system aligns with Vygotsky’s Zone of Proximal Development by recommending questions that are appropriately challenging—avoiding both boredom and frustration. This supports personalized learning, optimizes engagement, and reduces math anxiety, enhancing overall learning outcomes.

While effective at identifying relative difficulty, the system does not yet explain the underlying causes of difficulty or account for partial credit and affective factors (e.g., motivation, fatigue). 

\textbf{Future work:} We will improve interpretability by incorporating intrinsic algebraic features, integrating multi-modal learner engagement data (e.g., keystroke patterns, cursor movement, eye-tracking, EEG), adopting advanced exploration strategies such as Thompson Sampling, and extending the framework beyond algebra to other domains for richer, context-aware adaptation.

In summary, this research introduces a novel, data-driven approach to educational AI, capable of enhancing both learner experience and assessment quality in next-generation tutoring systems.


\phantomsection

\phantomsection
\section*{Funding}
\addcontentsline{toc}{section}{Funding}
This research received no specific grant from any funding agency in the public, commercial, or not-for-profit sectors..

\phantomsection
\section*{Declaration of Competing Interest}
\addcontentsline{toc}{section}{Declaration of Competing Interest}
The authors declare that they have no known competing financial interests or personal relationships that could have appeared to influence the work reported in this paper.\footnote{This publication is a preprint and has not been peer-reviewed. It is made available under the selected license. Any unauthorized use or violation of the license terms is strictly prohibited and will be the responsibility of the individual or entity engaging in such use.}

\phantomsection
\section*{Ethics Statement}
\addcontentsline{toc}{section}{Ethics Statement}
This study did not involve human participants or animals.

\phantomsection
\section*{Author Contributions}
\addcontentsline{toc}{section}{Author Contributions}
S.D. designed the study, analyzed the data, implemented the model, drafted the manuscript, and supervised the work. G.R. drafted literature review section. A.E. and R.K.R.  reviewed the manuscript. All authors read and approved the final version.



\begin{thebibliography}{00}

\bibitem{Audibert2009}
Audibert, J.-Y., Munos, R., Szepesvári, C.: Exploration-exploitation tradeoff using variance estimates in multi-armed bandits. Theor. Comput. Sci. \textbf{410}(19), 1876--1902 (2009). https://doi.org/10.1016/j.tcs.2009.01.016

\bibitem{Vermorel2005}
Vermorel, J., Mohri, M.: Multi-armed bandit algorithms and empirical evaluation. In: European Conference on Machine Learning, pp. 437--448. Springer (2005). https://doi.org/10.1007/11564096\_42

\bibitem{Ashcraft2002}
Ashcraft, M.H.: Math anxiety: Personal, educational, and cognitive consequences. Curr. Dir. Psychol. Sci. \textbf{11}(5), 181--185 (2002). https://doi.org/10.1111/1467-8721.00196

\bibitem{15}
Wang, Z., Lukowski, S.L., Hart, S.A., Lyons, I.M., Thompson, L.A., Kovas, Y., Mazzocco, M.M.M., Plomin, R., Petrill, S.A.: Is math anxiety always bad for math learning? The role of math motivation. Psychol. Sci. \textbf{26}(12), 1863--1876 (2015). https://doi.org/10.1177/0956797615602471

\bibitem{alkhuzaey2023}
AlKhuzaey, S. et al.: Text-based question difficulty prediction: A systematic review of automatic approaches. Int. J. Artif. Intell. Educ. \textbf{34}, 1--53 (2023). https://doi.org/10.1007/s40593-023-00362-1

\bibitem{kim2023}
Kim, G.I., Kim, S., Jang, B.: Classification of mathematical test questions using machine learning on datasets of learning management system questions. PLoS ONE \textbf{18}(10), e0286989 (2023). https://doi.org/10.1371/journal.pone.0286989

\bibitem{lin2023}
Lin, C.-C., Huang, A.Y.Q., Lu, O.H.T.: Artificial intelligence in intelligent tutoring systems toward sustainable education: a systematic review. Smart Learn. Environ. \textbf{10}(1), 41 (2023). https://doi.org/10.1186/s40561-023-00260-y

\bibitem{walkington2019supporting}
Walkington, C., Bernacki, M.L.: Personalizing algebra to students’ individual interests in an intelligent tutoring system: Moderators of impact. Int. J. Artif. Intell. Educ. \textbf{29}, 58--88 (2019). https://doi/10.1007/s40593-018-0168-1

\bibitem{Korhan2009}
Günel, K., Asliyan, R.: Determining difficulty of questions in intelligent tutoring systems. Turk. Online J. Educ. Technol. \textbf{8}, 14--21 (2009)

\bibitem{costa2023}
Costa, D.R., Chen, C.-W.: Exploring the relationship between process data and contextual variables among Scandinavian students on PISA 2012 mathematics tasks. Large-scale Assess. Educ. \textbf{11}(1), 5 (2023). https://doi.org/10.1186/s40536-023-00155-x

\bibitem{albano2020}
Albano, A.D., McConnell, S.R., Lease, E.M., Cai, L.: Contextual interference effects in early assessment: Evaluating the psychometric benefits of item interleaving. Front. Educ. \textbf{5}, 133 (2020). https://doi.org/10.3389/feduc.2020.00133

\bibitem{anderson2014}
Anderson, J.R., Lee, H.S., Fincham, J.M.: Discovering the structure of mathematical problem solving. NeuroImage \textbf{97}, 163--177 (2014). https://doi.org/10.1016/j.neuroimage.2014.04.031

\bibitem{hong2021}
Hong, Y.-S., Han, C.-P., Cho, S.-S.: Level-based learning algorithm based on the difficulty level of the test problem. Appl. Sci. \textbf{11}(10), 4380 (2021). https://doi.org/10.3390/app11104380

\bibitem{Pado2017}
Padó, U.: Question difficulty--how to estimate without norming, how to use for automated grading. In: Proceedings of the 12th Workshop on Innovative Use of NLP for Building Educational Applications, pp. 1--10 (2017). https://doi.org/10.18653/v1/W17-5001

\bibitem{lee2000}
Lee, F.-L., Heyworth, R.: Problem complexity: A measure of problem difficulty in algebra by using computer. Educ. J. \textbf{28}(1), 85--108 (2000)

\bibitem{chen2005}
Chen, C.-M., Lee, H.-M., Chen, Y.-H.: Personalized e-learning system using item response theory. Comput. Educ. \textbf{44}(3), 237--255 (2005). https://doi.org/10.1016/j.compedu.2004.01.006

\bibitem{Segal2018}
Segal, A., Ben David, Y., Williams, J.J., Gal, K., Shalom, Y.: Combining difficulty ranking with multi-armed bandits to sequence educational content. In: Artificial Intelligence in Education, pp. 317--321. Springer (2018). https://doi.org/10.1007/978-3-319-93846-2\_59

\bibitem{Jiao2023}
Jiao, Y., Shridhar, K., Cui, P., Zhou, W., Sachan, M.: Automatic educational question generation with difficulty level controls. In: International Conference on Artificial Intelligence in Education, pp. 476--488. Springer (2023). https://doi.org/10.1007/978-3-031-36272-9\_39

\bibitem{Ningsih2023}
Ningsih, S.R., Romadhony, A.: Difficulty level identification of Indonesian and mathematics multiple choice questions using machine learning approach. Build. Inform. Technol. Sci. \textbf{5}(1), 236--245 (2023). https://doi.org/10.47065/bits.v5i1.3649

\bibitem{Shabana2023}
Shabana, K.M., Lakshminarayanan, C.: Unsupervised concept tagging of mathematical questions from student explanations. In: International Conference on Artificial Intelligence in Education, pp. 627--638. Springer (2023). https://doi.org/10.1007/978-3-031-36272-9\_51

\bibitem{Fernandez-Fontelo2023}
Fernández-Fontelo, A., Kieslich, P.J., Henninger, F., Kreuter, F., Greven, S.: Predicting question difficulty in web surveys: A machine learning approach based on mouse movement features. Soc. Sci. Comput. Rev. \textbf{41}(1), 141--162 (2023). https://doi.org/10.1177/08944393211032950

\bibitem{Nurhalimah2022}
Nurhalimah, A., Mandailina, V., Mahsup, Syaharuddin: Measuring the difficulty level of mathematical problems based on Polya criteria. J. Educ. Res. Eval. \textbf{6}(3), 595--607 (2022). https://doi.org/10.23887/jere.v6i4.46316

\bibitem{kinshuk2002}
Kinshuk: Does intelligent tutoring have future!. In: International Conference on Computers in Education, 2002. Proceedings., pp. 1524--1525. IEEE (2002). https://doi.org/10.1109/CIE.2002.1186328

\bibitem{ram2024}
Ram, I., Harris, S., Roll, I.: Choice-based personalization in MOOCs: Impact on activity and perceived value. Int. J. Artif. Intell. Educ. \textbf{34}, 376--394 (2024). https://doi.org/10.1007/s40593-023-00334-5

\bibitem{brooks2021}
Brooks, C., Quintana, R.M., Choi, H., Quintana, C., NeCamp, T., Gardner, J.: Towards culturally relevant personalization at scale: Experiments with data science learners. Int. J. Artif. Intell. Educ. \textbf{31}, 516--537 (2021). https://doi.org/10.1007/s40593-021-00262-2

\bibitem{roll2018}
Roll, I., Russell, D.M., Gašević, D.: Learning at scale. Int. J. Artif. Intell. Educ. \textbf{28}(4), 471--477 (2018). https://doi.org/10.1007/s40593-018-0170-7

\bibitem{zhu2021}
Zhu, M., Sabır, N., Bonk, C.J., Sarı, A., Xu, S., Kım, M.: Addressing learner cultural diversity in MOOC design and delivery: Strategies and practices of experts. Turk. Online J. Distance Educ. \textbf{22}(2), 1--25 (2021). https://doi.org/10.17718/tojde.906468

\bibitem{liu2016}
Liu, Z., Brown, R., Lynch, C.F., Barnes, T., Baker, R., Bergner, Y., McNamara, D.: MOOC learner behaviors by country and culture; An exploratory analysis. In: Proceedings of the 9th International Conference on Educational Data Mining, pp. 127--134 (2016)

\bibitem{wang2007}
Wang, M.: Designing online courses that effectively engage learners from diverse cultural backgrounds. Br. J. Educ. Technol. \textbf{38}(2), 294--311 (2007). https://doi.org/10.1111/j.1467-8535.2006.00626.x

\bibitem{elsabagh2021}
El-Sabagh, H.A.: Adaptive e-learning environment based on learning styles and its impact on development students' engagement. Int. J. Educ. Technol. High. Educ. \textbf{18}, 53 (2021). https://doi.org/10.1186/s41239-021-00289-4

\bibitem{mirata2020}
Mirata, V., Hirt, F., Bergamin, P., Ziska, S.: Challenges and contexts in establishing adaptive learning in higher education: findings from a Delphi study. Int. J. Educ. Technol. High. Educ. \textbf{17}, 32 (2020). https://doi.org/10.1186/s41239-020-00209-y

\bibitem{Molina2019}
Molina del Río, J., Guevara, M.A., Hernández González, M., Harmony, T., Fernández-Bouzas, A., Galán, L., Valdés-Sosa, P.: EEG correlation during the solving of simple and complex logical–mathematical problems. Cogn. Affect. Behav. Neurosci. \textbf{19}(4), 1036--1046 (2019). https://doi.org/10.3758/s13415-019-00703-5

\bibitem{pomerantz2005}
Pomerantz, J.: A linguistic analysis of question taxonomies. J. Am. Soc. Inf. Sci. Technol. \textbf{56}(7), 715--728 (2005). https://doi.org/10.1002/asi.20162

\bibitem{moyer2021}
Moyer, M.C., Syrett, K.: The study of questions. Wiley Interdiscip. Rev. Cogn. Sci. \textbf{12}(1), e1531 (2021). https://doi.org/10.1002/wcs.1531

\bibitem{beymer2024students}
Beymer, P.N., Schell, M.J., Alberts, K.M., Phun, V., Rosenberg, J.M., Schmidt, J.A.: Students' situational engagement profiles in formal and informal science learning environments. J. Res. Sci. Teach. (2024). https://doi.org/10.1002/tea.22017

\bibitem{payne2019}
Payne, L.: Student engagement: Three models for its investigation. J. Further High. Educ. \textbf{43}(5), 641--657 (2019). https://doi.org/10.1080/0309877X.2017.1391186

\bibitem{bryson2010}
Bryson, C., Hardy, C.: Reaching a common understanding of the meaning of student engagement. In: Society for Research into Higher Education Annual Research Conference 2010, Wales (2010)

\bibitem{sheard2013difficult}
Sheard, J., Carbone, A., Chinn, D., Clear, T., Corney, M., D'Souza, D., Fenwick, J., Harland, J., Laakso, M.-J., Teague, D., et al.: How difficult are exams? A framework for assessing the complexity of introductory programming exams. In: Proceedings of the 15th Australasian Computing Education Conference, pp. 145--154. Australian Computer Society (2013). https://doi.org/10.5555/2667199.2667215

\bibitem{he2023question}
He, J., Chen, J., Peng, L., Sun, B., Zhang, H.: Question difficulty prediction with external knowledge. In: Proceedings of the 2023 12th International Conference on Software and Computer Applications, pp. 59--64 (2023). https://doi.org/10.1145/3587828.3587838

\bibitem{matsuda2019mathematical}
Matsuda, T., Sonoda, M., Eto, M., Satoh, H., hanada, T., Kanahama, N., Katoh, D., Ishikawa, H.: Mathematical question structure extraction and possibilities of automatic question making by plane graph. In: Proceedings of the 2019 3rd International Conference on Education and E-Learning, pp. 109--112 (2019). https://doi.org/10.1145/3371647.3371655

\bibitem{sujay2024multi}
Sujay, R., Perumal, S., Nagraj, Y., Ghei, A., Srinivas, K.S.: Multi-faceted question complexity estimation targeting topic domain-specificity. In: CS \& IT Conference Proceedings, vol. 14 (2024). https://doi.org/10.5121/csit.2024.141513

\bibitem{chrysostomou2019analysing}
Chrysostomou, M.-B., Christou, C.: Analysing the notion of algebraic thinking based on empirical evidence/Un análisis del concepto de pensamiento algebraico basado en evidencia empírica. J. Study Educ. Dev. \textbf{42}(3), 721--781 (2019). https://doi.org/10.1080/02103702.2019.1604022

\bibitem{watkins1989learning}
Watkins, C.J.C.H.: Learning from Delayed Rewards. Ph.D. thesis, King's College, University of Cambridge (1989)

\bibitem{mahajan2008multi}
Mahajan, A., Teneketzis, D.: Multi-armed bandit problems. In: Foundations and Applications of Sensor Management, pp. 121--151. Springer (2008). https://doi.org/10.1007/978-0-387-49819-5\_6

\bibitem{vakili2013deterministic}
Vakili, S., Liu, K., Zhao, Q.: Deterministic sequencing of exploration and exploitation for multi-armed bandit problems. IEEE J. Sel. Top. Signal Process. \textbf{7}(5), 759--767 (2013). https://doi.org/10.1109/JSTSP.2013.2263494

\bibitem{carpentier2011ucb}
Carpentier, A., Lazaric, A., Ghavamzadeh, M., Munos, R., Auer, P.: Upper-confidence-bound algorithms for active learning in multi-armed bandits. In: Algorithmic Learning Theory, pp. 318--332. Springer (2011). https://doi.org/10.1007/978-3-642-24412-4\_17

\bibitem{liu2020risk}
Liu, X., Derakhshani, M., Lambotharan, S., Van der Schaar, M.: Risk-aware multi-armed bandits with refined upper confidence bounds. IEEE Signal Process. Lett. \textbf{28}, 269--273 (2020). https://doi.org/10.1109/LSP.2020.3047725

\bibitem{agrawal2012analysis}
Agrawal, S., Goyal, N.: Analysis of thompson sampling for the multi-armed bandit problem. In: Conference on Learning Theory, pp. 39--1. JMLR Workshop and Conference Proceedings (2012)

\bibitem{roijers2017interactive}
Roijers, D.M., Zintgraf, L.M., Nowé, A.: Interactive thompson sampling for multi-objective multi-armed bandits. In: International Conference on Algorithmic Decision Theory, pp. 18--34. Springer (2017). https://doi.org/10.1007/978-3-319-67504-6\_2

\bibitem{hardwick1991bandit}
Hardwick, J., Stout, Q.F., et al.: Bandit strategies for ethical sequential allocation. Comput. Sci. Stat. \textbf{23}(6.1), 421--424 (1991)

\bibitem{awerbuch2004adaptive}
Awerbuch, B., Kleinberg, R.D.: Adaptive routing with end-to-end feedback: Distributed learning and geometric approaches. In: Proceedings of the Thirty-Sixth Annual ACM Symposium on Theory of Computing, pp. 45--53 (2004). https://doi.org/10.1145/1007352.1007361

\bibitem{orsoni2023recommending}
Orsoni, M., Pögelt, A., Duong-Trung, N., Benassi, M., Kravcik, M., Grüttmüller, M.: Recommending mathematical tasks based on reinforcement learning and item response theory. In: Augmented Intelligence and Intelligent Tutoring Systems, pp. 17--28. Springer (2023). https://doi.org/10.1007/978-3-031-32883-1\_2

\bibitem{murray2002toward}
Murray, T., Arroyo, I.: Toward measuring and maintaining the zone of proximal development in adaptive instructional systems. In: Intelligent Tutoring Systems, pp. 749--758. Springer (2002). https://doi.org/10.1007/3-540-47987-2\_75

\bibitem{vygotsky1978mind}
Vygotsky, L.S.: Mind in Society. Harvard University Press, Cambridge, MA (1978)

\end{thebibliography}
\end{document}